# Effect of Reporting Mode and Clinical Experience on Radiologists' Gaze and Image Analysis Behavior in Chest Radiography

Mahta Khoobi[*,1,2], Marc S. von der Stueck[2], Felix Barajas Ordonez[1,2], Anca-Maria Iancu[2], Eric N. Corban[2], Julia Nowak[2], Aleksandar Kargaliev[2], Valeria Perelygina[2], Anna-Sophie Schott[2], Daniel Pinto dos Santos[3], Christiane Kuhl[2], Daniel Truhn[1,2], Sven Nebelung[1,2], Robert M. Siepmann[1,2]

[1]Lab for Artificial Intelligence in Medicine, Department of Diagnostic and Interventional Radiology, University Hospital Aachen, Aachen, Germany
[2]Department of Diagnostic and Interventional Radiology, University Hospital Aachen, Aachen, Germany
[3]Department of Diagnostic and Interventional Radiology, University Medical Center of Johannes Gutenberg-University Mainz, Mainz, Germany

## Abstract

**Background:** Structured reporting (SR) and artificial intelligence (AI) potentially transform how radiologists interact with imaging studies.

**Purpose:** To assess the impact of different reporting modes (with or without AI) on image analysis behavior, diagnostic accuracy and efficiency, and user experience.

**Materials and Methods:** In this prospective study (July to December 2024), we recruited four novice and four non-novice readers (radiologists and pre-graduate medical students). Each reader analyzed 35 bedside chest radiographs per session using one of three reporting modes: free-text (FT) reporting, SR with itemized and graded findings, and SR with AI-prefilled suggestions (AI-SR). A customized viewer presented the radiographs and reporting interface on a screen-based eye-tracking system. Outcome measures included diagnostic accuracy (compared to the majority vote of expert radiologists, quantified using Cohen's κ coefficient), reporting time per radiograph, eye-tracking metrics, and questionnaire-based user experience. Statistical analyses were performed using generalized linear mixed models and Bonferroni post-hoc tests, with a significance level of $P \leq .01$.

**Results:** Diagnostic accuracy was similar in FT (κ=0.58) and SR (κ=0.60) but significantly higher in AI-SR (κ=0.71) ($P<.001$). SR and AI-SR significantly reduced reporting times per radiograph from 88±38 seconds (FT) to 37±18 seconds (SR) and 25±9 seconds (AI-SR) ($P<.001$). Saccade counts (quick eye movements between fixations) for the radiograph display field (205±135 [FT], 123±88 [SR], 97±58 [AI-SR]) and total fixation duration for the report display field (11±5 seconds [FT], 5±3 seconds [SR], 4±1 seconds [AI-SR]) significantly decreased with SR and AI-SR ($P<.001$ each). For novice readers, the gaze focus shifted from the report to the radiograph display field in SR, while non-novice readers maintained their visual focus on the radiograph, irrespective of the reporting mode. AI-SR was the preferred reporting mode.

**Conclusion:** Compared to free-text reporting, structured reporting enhances efficiency by guiding visual attention toward the image, and AI-prefilled structured reporting improves diagnostic accuracy.

**Corresponding Author***

**Mahta Khoobi, M. Sc.**
Lab for Artificial Intelligence in Medicine
Department of Diagnostic and Interventional Radiology
University Hospital RWTH Aachen
Pauwelsstr. 30
52074 Aachen, Germany
Email: mkhoobi@ukaachen.de

# 1. Introduction

Radiology has recently experienced major transformative changes, notably the introduction of structured reporting (SR) and the advent of artificial intelligence (AI).

The clinical impact of SR is debated. On the one hand, it increases linguistic standardization (1) and improves reporting completeness, clinical decision-making, and information extraction compared to traditional free-text (FT) reports (2). On the other hand, SR may divert a radiologist's focus from the diagnostic task toward the reporting system (3), and its rigid reporting templates may fail to capture the nuances of complex or atypical findings (4, 5).

Similarly, the clinical impact of AI is an area of active research. AI may assist radiologists by highlighting potentially overlooked findings (6) and reducing their workload (7). Despite these benefits, the clinical adoption of AI faces numerous practical challenges (8) and raises concerns about automation bias and overreliance, particularly among less experienced radiologists (9). Moreover, whether AI assistance affects a radiologist's image analysis behavior is unclear.

While FDA- and CE-approved AI tools for chest radiography are increasingly available (8), their impact on the diagnostic process remains underexplored. Anecdotal evidence suggests that AI-prefilled structured reporting (AI-SR) improves reporting quality and efficiency (10).

With the growing demand for radiologic services and increasing imaging volumes and complexity (11), it is critical to develop reporting workflows that balance efficiency and accuracy while minimizing distractions and reducing non-image interactions. However, little is known about how different reporting modes –and AI assistance– impact diagnostic workflows. To fill this gap, we aimed to evaluate the effects of various reporting modes on radiologists' interactions with chest radiographs and their diagnostic accuracy, efficiency, and user experience. We hypothesized that different reporting modes would affect image interaction and diagnostic efficiency but not diagnostic accuracy.

## 2. Materials & Methods

### 2.1 Study Design

The study was designed as a prospective, intra-individual comparative reader study to evaluate the effects of free-text reporting, structured reporting, and the use of an AI model developed and validated on a large retrospective chest radiograph dataset on the diagnostic workflow (**Figure 1**). This study has been approved by the Institutional Review Board of the Medical Faculty of RWTH Aachen University under protocol number EK24-244.

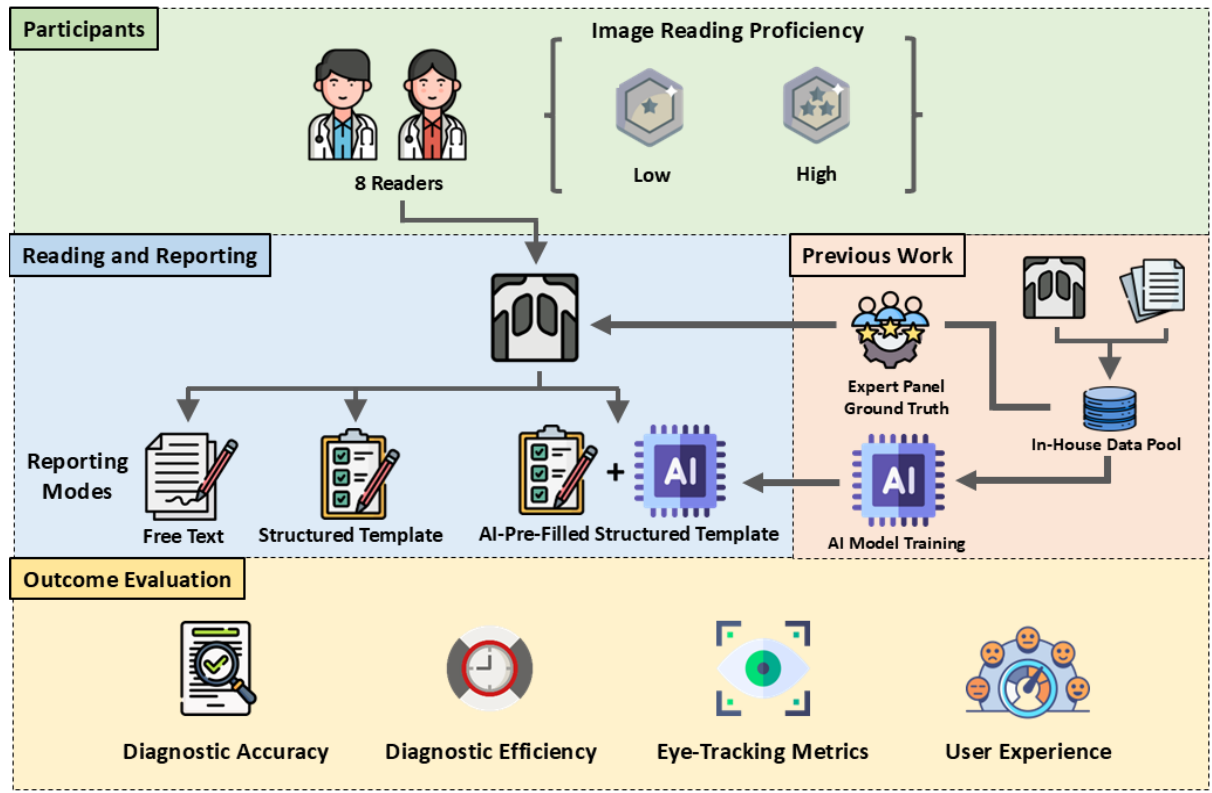

**Figure 1: Study Design and Workflow.** Readers were assigned to two levels based on their experience in reading radiographs: pre-graduate medical students and radiologists without formal training in radiography (novice group) vs. radiologists with completed training in radiography but pending board certification (non-novice group). After reading 35 chest radiographs per session in three sessions across three reporting modes, separated by two-week washout periods, outcomes were evaluated along the indicated dimensions. The in-house data pool of 193,566 radiographs (from 45,016 patients) and associated radiologic reports were used for training the Artificial Intelligence (AI) model. Of the held-out internal test set, 35 radiographs from 35 patients were assessed by an expert panel of six radiologists, considered the ground truth, and included in the present study.

### 2.2 Dataset Characteristics

This study is based on our previously published in-house dataset of 193,556 bedside chest radiographs from 45,016 patients acquired between January 2009 and December 2020 (12), and treated in one of our tertiary academic medical center's (University Hospital Aachen, Germany) intensive care units. The radiographs had been ordered by specialty-trained physicians for various clinical indications. A representative set of 35 chest radiographs of 35 unique patients (age: 70.2±17.1 years, 19 females) was selected in a stratified manner to contain a clinically realistic distribution of variable imaging findings (**Supplementary Table 1**). Consequently, radiographs with mostly mild and occasionally moderate imaging findings as well as normal radiographs were included. The radiographs were independently assessed by six expert radiologists (mean clinical experience: 6.8 years) based on a structured, itemized, and semiquantitative reporting template indicating the absence (or presence) and the severity of cardiomegaly, pulmonary congestion, pleural effusion (left/right), pulmonary opacities (left/right), and atelectasis (left/right). Here, 'none' indicates absence; (+) questionable presence; '+' mild presence (regarding distribution and severity); '++' moderate presence; and '+++' severe presence. The majority vote of the six expert radiologists provided the ground truth.

AI predictions were generated using a standard convolutional neural network trained on the above dataset (training set, n=122,294 radiographs; validation set, n=31,243 radiographs), with very good agreement with the majority vote of six expert radiologists (Cohen's κ=0.86; **Supplementary Text 1**) (12). Following anonymization, the radiographs were exported as DICOM images.

Assuming a moderate effect size, we determined a minimum sample size of 34 radiographs per session. A detailed power analysis is provided in **Supplementary Text 2**.

## 2.3 Experimental Setup

### 2.3.1 Reader Selection

While the majority vote of six expert radiologists provided the reference standard for the radiographs, six clinical resident radiologists with varying experience levels and two pre-graduate medical students were recruited for the reader study and assigned one of two experience levels based on their experience in reading radiographs. In our institution, pre-graduate medical students during their 4-month radiology rotation in their final year also report radiographs under supervision, similar to first-year residents. Hence, the novice group consisted of four readers (three female): two pre-graduate medical students and two residents without complete training in radiography with a mean clinical experience of 6.8 (±4.8 [Standard Deviation]) months (range, 3 to 15 months), during which time they had reported (under supervision) a mean of 75 (±110) chest radiographs (range, 0 to 264). The non-novice group consisted of four residents with complete training in radiography but pending board certification (one female). Their clinical experience was 43.5 (±10.3) months (range, 26 to 52 months), during which time they had reported 4,523 (±1,885) chest radiographs (range, 1,789 to 6,561).

This study was approved by the local ethical committee (Ethical Committee, Medical Faculty, RWTH Aachen University, reference number 244/24). Informed consent was obtained from each reader in written form.

### 2.3.2 Technical Setup

**Figure 2** illustrates the technical setup. The study was conducted in a simulated radiologic reading room with dimmable lighting, blackout windows, and low noise levels. Radiographs were read using a customized DICOM viewer based on the Open Health Imaging Foundation platform (13), with images served via a local DICOMWeb API using the Orthanc server (14). The viewer displayed the radiograph on the right side of the screen ("radiograph display field") and the reporting interface on the left ("report display field"). Eye movements were tracked using a 24-inch monitor with an embedded Tobii TX300 eye-tracking device (15), capable of binocular tracking at 300 Hz. Calibration was performed before each session using Tobii Studio software (v3.3). Eye motion was continuously recorded, and only readings with at least 60% correctly sampled gaze data were included.

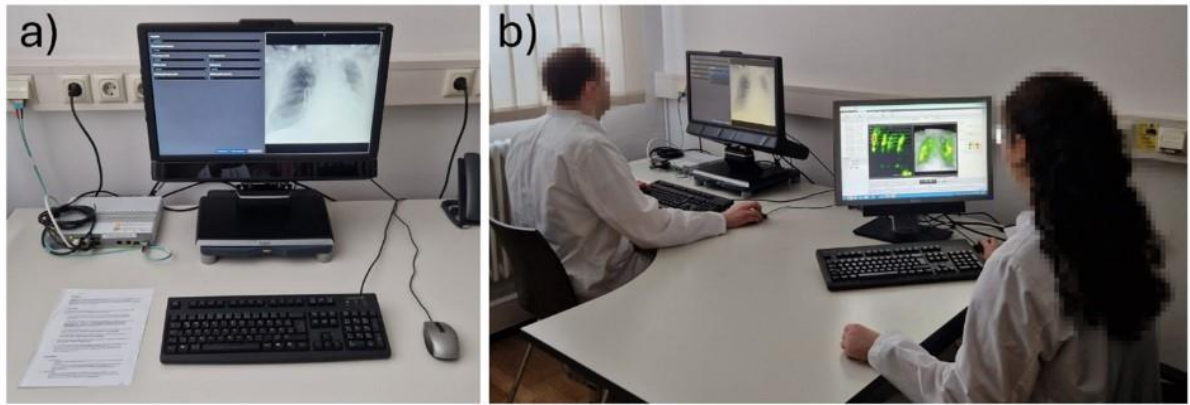

**Figure 2: Technical Setup for Reporting Under Eye-Tracking Conditions.** a) On a dedicated screen-based eye-tracking device, anteroposterior chest radiographs (right screen half [radiograph display field]) were read and reported upon (left screen half [report display field]). Here, the itemized template for structured reporting of graded imaging findings is shown. The paper sheet next to the keyboard detailed the study instructions. b) On the associated workstation (foreground), the captured eye-tracking data were registered and saved for processing.

### 2.3.3 Reporting Modes

Before calibration, the study instructions were presented, which included a general introduction and an explanation of the different reporting modes and procedures. The AI model was framed explicitly as a 'work in progress' that demonstrates agreement with expert radiologists but may make mistakes.

The readers were asked to read and report the chest radiographs in three reporting modes (**Figure 3**):

i. **Free-Text Reporting (FT):** Readers phrased the reports in their own words. Readers were asked to report the presence and severity/distribution of the must-report imaging findings detailed above. The text expansion and automation software (AutoHotkey, v.2.0.18) was enabled to assist the readers in streamlining the production of frequently used text snippets by customized keyboard shortcuts.
ii. **Structured Reporting (SR):** Readers were asked to fill an empty structured template based on itemized and graded imaging findings to indicate their presence and severity. This template was and is used for clinical reporting of chest radiographs at our institution.
iii. **AI-Prefilled Structured Reporting (AI-SR):** The SR template (ii) was prefilled with the AI model's predictions. Readers could modify or adopt the suggestions.

During reporting, the radiographs could not be zoomed or moved to ensure that eye-tracking measurements were not affected by fixations related to these functionalities. A board-certified senior radiologist (S.N., nine years of clinical experience) ensured that all relevant imaging findings could be established without image manipulation. Following their submission, reports could not be revisited or altered. Reports were saved alongside the patient identifier and timestamps. Per session, each reader analyzed 38 radiographs. The first three radiographs were intended to familiarize the readers with the setup and were disregarded. Consequently, each reader's interaction with 35 consecutive radiographs was analyzed. Readers provided feedback on user experience at the end of each reading session. A washout period of at least two weeks was maintained between the reading sessions, and the same radiographs were administered in a different order.

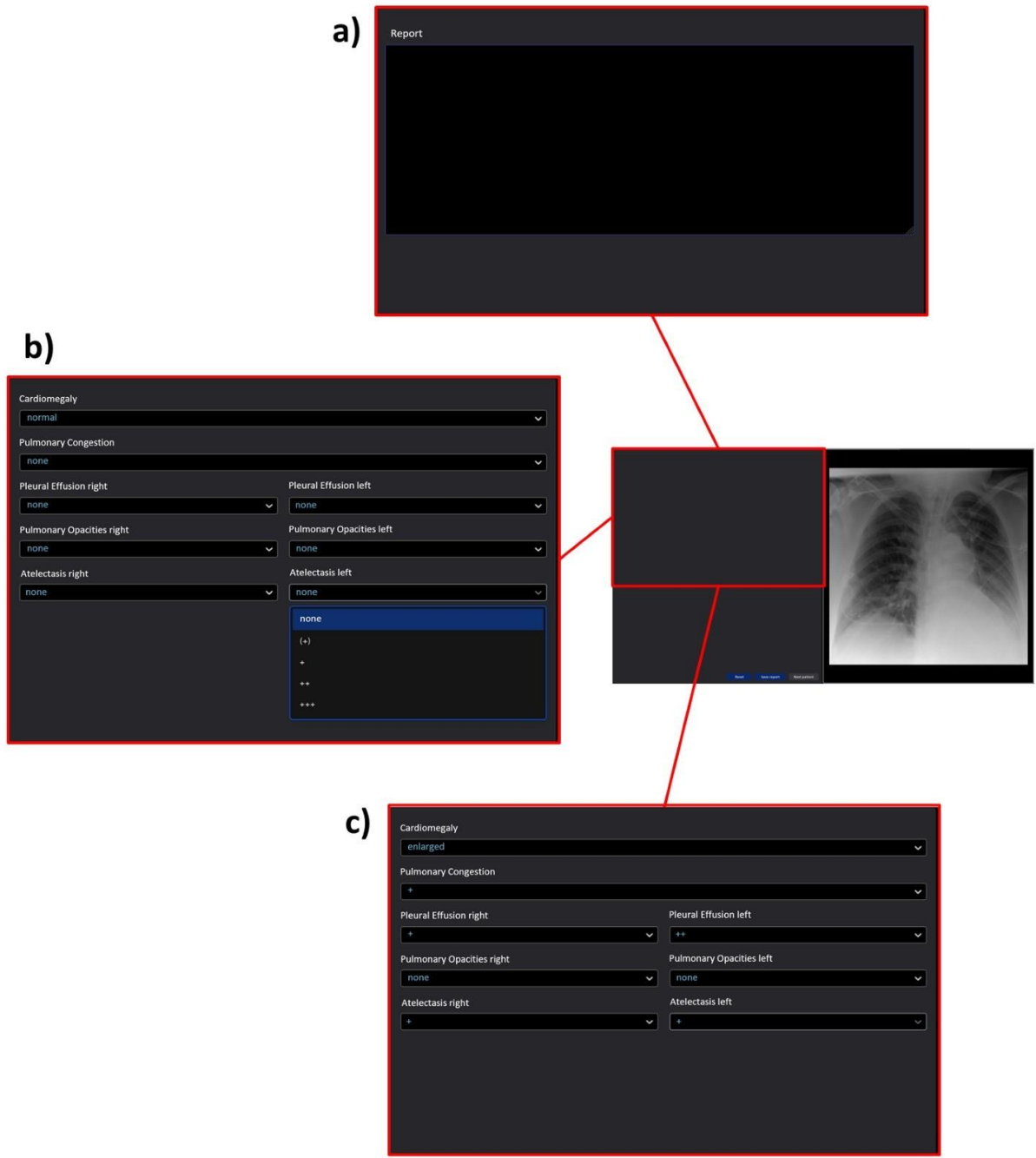

**Figure 3: Visualization of Reporting Modes Alongside the Radiograph for Eye Tracking.** The DICOM viewer was customized to visualize the chest radiograph on the right half of the screen (radiograph display field) and the reporting interface on the left (report display field). Larger red boxes are zoomed-in versions of the small red box. a) A text box covering 25% of the screen was used for free text reports. b) A template was used for structured reporting and could be altered using a drop-down list for each imaging finding to indicate its presence and severity/distribution. By default, "none" or "normal" (for cardiomegaly) was set. c) The same template was prefilled with the predictions of the Artificial Intelligence model.

## 2.4. Outcome Metrics

**Diagnostic accuracy** was evaluated by comparing the reported imaging findings with the ground truth provided by the majority vote of six expert radiologists. For the free-text mode, reported imaging findings were categorized based on presence or absence and severity/distribution. To this end, imaging findings referred to as 'mild' or 'weak' were graded as '+,' while 'moderate' or 'modest' were graded as '++,' and 'severe' or 'marked' were graded as '+++.' Graded imaging findings were available for the structured reporting and AI-prefilled structured reporting modes.

Average **diagnostic efficiency** was quantified by recording the total time needed to report on all radiographs per reading session and dividing by the number of radiographs, yielding the average reporting time per radiograph.

**Eye-tracking metrics** were parameterized and quantified using Tobii Studio software, except for the saccade count, which was computed based on the raw data (**Supplementary Text 3**). After manually segmenting time frames corresponding to the reader's interactions with one radiograph, two areas of interest (AOI) were defined as the reporting interface (report display field, $AOI_1$) and the radiograph (radiograph display field, $AOI_2$). The eye-tracking metrics (**Table 1**) were calculated for $AOI_1$, $AOI_2$, and their combination.

**User experience feedback** was collected on a per-reader and per-session basis using surveys. Readers were asked to rate diagnostic confidence, user satisfaction, cognitive ease, user interface (convenience, efficiency), clinical adoption willingness, usefulness, and trust in AI assistance as low, neutral, or high (**Supplementary Table 2**).

**Table 1: Definitions of Acquired Eye-Tracking Metrics.**

| Metric [Unit] | Definition |
| --- | --- |
| Fixation | A period during which the eyes remain stationary, focusing on a specific AOI. |
| Fixation Count [n] | Total number of fixations within an AOI. |
| Total Fixation Duration [s] | Cumulative duration of all fixations within an AOI. |
| Visit | Sequence of fixations and saccades within an AOI without any intervening fixations outside that AOI. |
| Visit Count [n] | Total number of visits to an AOI. |
| Total Visit Duration [s] | Cumulative duration of all visits to an AOI. |
| Saccade | Rapid eye movement between fixations to reposition the fovea to a new location in the visual field for detailed examination. |
| Saccade Count [n] | Total number of saccades per AOI. |

Note: AOI – Area of Interest, i.e., a specific region defined for analysis.

## 2.5. Statistical Analysis

Diagnostic accuracy was quantified using the quadratic weighted Cohen's κ coefficient, calculated using Python (v3.12) and its libraries NumPy (v2.1.0) and SciPy (v1.14.1). Additionally, confusion-matrix-based metrics such as true/false positives and true/false negatives, and sensitivity, specificity, and accuracy were calculated as a function of reporting mode and reader experience.

κ coefficients, reporting times per radiograph, and eye-tracking metrics were analyzed using generalized linear mixed models implemented in R (v4.4.2, R Foundation for Statistical Computing, Vienna, Austria) (16) using RStudio (v2024.09.0, RStudio, Boston, MA, USA) (17). Reporting mode, reader experience level, and AOI were considered fixed predictors, while the reader was considered the random effect. For κ coefficients, a beta distribution and logit link function (18) were applied. Natural log transformation was applied to positively skewed data for more robust statistical modeling. Model assumptions were confirmed using tests for residual distributions, variance consistency, predictor correlations, and overall model fit, complemented by QQ plots and Shapiro-Wilk tests. The

Type III Wald Chi-square test was used to assess the potential significance of predictors. Following significant main effects, pairwise post-hoc comparisons were conducted using estimated marginal means derived from fitted models within experience levels (all outcome metrics) and AOIs (eye-tracking metrics). Pairwise tests were adjusted using the Bonferroni method, accounting for three comparisons (i.e., reporting modes). P-values of ≤0.01 were considered statistically significant to reduce the number of statistically significant yet clinically likely irrelevant findings. Data are means ± standard deviation.

## 3. Results

The study was conducted between July and December 2024, and all readers completed the three reading sessions.

A summary of the key findings across reporting modes is provided in **Table 2**.

**Diagnostic accuracy** varied with reporting mode (**Table 3**). Free-text and structured reporting showed comparable agreement with the reference standard (κ=0.54-0.66), whereas AI-prefilled structured reporting improved accuracy to κ=0.71 (P<.001). The neural network outperformed all readers and conditions with κ=0.81, reflecting critical reader interaction with the pre-filled template. For all readers, accuracy derived from confusion matrices improved from 77% (free-text reporting) to 82% (AI-prefilled structured reporting), driven mainly by gains in sensitivity (**Supplementary Table 3**). Novice readers showed the largest improvements in accuracy, whereas experienced readers maintained stable performance across reporting modes.

Average **diagnostic efficiency** was significantly higher for structured reporting than for free-text reporting (all experience levels) and for AI-prefilled structured reporting than for structured reporting (novice readers only) (P≤.001; **Table 3**). Novice readers experienced the largest improvement in efficiency, with significant reductions in reporting time with structured reporting and AI-prefilled structured reporting. While non-novice readers also showed decreased reporting times with structured reporting and AI-prefilled structured reporting, the additional efficiency gains from AI-prefilled structured reporting versus structured reporting were not significant (P=.125).

**Image analysis behavior**, as assessed by eye-tracking metrics, was impacted by reporting mode, reader experience, and AOI (**Table 4, Supplementary Table 4** [post-hoc details]). In line with the findings above, visual processing became more efficient for all readers using structured reporting and AI-prefilled structured reporting (versus free-text reporting), as indicated by significantly decreased fixation durations and saccade counts. When considered alongside other visit-associated metrics, efficiency gains resulted from reductions in time and eye movements required during reporting. Also, the cumulative visual attention, as assessed by total fixation duration and total visit duration, was shifted from the report to the radiograph display field. Relative decreases were larger for the report than the radiograph display field (**Supplementary Table 5**). Novice readers benefitted most from structured reporting and AI-prefilled structured reporting, as indicated by significantly longer interaction with the radiograph than the report display field – both for total duration fixation and total visit duration. For non-novice readers, significant reductions were observed for the report display field when using structured reporting and AI-prefilled structured reporting. For the radiograph display field, these changes were not significant (e.g., P=.506 [total fixation duration]), and accordingly, no significant differences were found between both visual fields for any metric. Notably, while metrics tended to be lowest for AI-prefilled structured reporting, post-hoc testing revealed no significant differences in eye-tracking metrics between structured reporting and AI-prefilled structured reporting, regardless of experience level or display field.

**Table 2: Key Findings at a Glance - Pairwise Differences Between Reporting Modes**

| Outcome (all readers) | SR vs FT | AI-SR vs FT | AI-SR vs SR |
|---|---|---|---|
| Diagnostic accuracy (κ) | — (Δ +0.02) | ▲ +0.13 * | ▲ +0.11 * |
| Reporting time / radiograph (s) | ▼ –51 s * | ▼ –63 s * | ▼ –12 s * |
| Total fixation duration (report display field) (s) | ▼ –6.6 s * | ▼ –7.8 s * | — (Δ –1.2 s) |
| Total fixation duration (radiograph display field) (s) | — (Δ –6.9 s) | — (Δ –10.6 s) | — (Δ –3.7s) |
| Fixation count (report display field) (n) | — (Δ –13.5) | ▼ –18.5 * | — (Δ 5.0) |
| Fixation count (radiograph display field) (n) | — (Δ –19.4) | — (Δ –38.4) | — (Δ –19.0) |
| Saccade count (report display field) (n) | ▼ –47.3 * | ▼ –45.8 * | — (Δ +1.5) |
| Saccade count (radiograph display field) (n) | — (Δ –82.0) | ▼ –108.2 * | — (Δ –26.2) |
| Total visit duration (report display field) (s) | ▼ –10.9 s * | ▼ –12.3 s * | — (Δ –1.4 s) |
| Total visit duration (radiograph display field) (s) | — (Δ –8.6 s) | ▼ –14.5 s * | — (Δ –5.9 s) |
| Visit count (report display field) (n) | — (Δ –1.9) | ▼ –3.8 * | — (Δ –1.9) |
| Visit count (radiograph display field) (n) | — (Δ –0.3) | — (Δ –3.2) | — (Δ –1.9) |

Note: Each cell shows the direction (▲ increase, ▼ decrease) and magnitude (absolute Δ) of the change for the indicated outcome, as well as a significance marker (*P <.01, Bonferroni-adjusted). "SR vs FT" compares structured reporting with free-text reporting; "AI-SR vs FT" compares AI-prefilled structured reporting with free-text reporting; "AI-SR vs SR" compares AI-prefilled structured reporting with structured reporting. Absolute values and full model statistics are reported in Tables 3 and 4 as well as in Supplementary Tables 3-6. Abbreviations: FT = free-text reporting; SR = structured reporting; AI-SR = AI-prefilled structured reporting.

**Table 3: Diagnostic Accuracy and Mean Reporting Time per Radiograph as a Function of Reporting Mode and Reader Experience.**

| Metric | Reader Experience | FT | SR | AI-SR | P-value | Post-hoc Details | | |
|---|---|---|---|---|---|---|---|---|
| | | | | | | FT vs. SR | FT vs. AI-SR | SR vs. AI-SR |
| **Diagnostic Accuracy** | All | 0.58 ± 0.06 | 0.60 ± 0.07 | 0.71 ± 0.03 | **<.001** | 1.0 | **<.001** | **<.001** |
| | Low | 0.54 ± 0.04 | 0.54 ± 0.06 | 0.71 ± 0.01 | **<.001** | 1.0 | **<.001** | **<.001** |
| | High | 0.62 ± 0.04 | 0.66 ± 0.02 | 0.71 ± 0.04 | **<.001** | 0.10 | **<.001** | **<.001** |
| **Mean Reporting Time per Radiograph** | All | 88.1 ± 38.4 | 37.3 ± 18.2 | 25.0 ± 8.8 | **<.001** | **<.001** | **<.001** | **<.001** |
| | Low | 114.0 ± 39.2 | 51.0 ± 16.0 | 31.1 ± 7.7 | **<.001** | **<.001** | **<.001** | **<.001** |
| | High | 62.2 ± 10.5 | 23.6 ± 4.6 | 18.8 ± 4.3 | **<.001** | **<.001** | **<.001** | .125 |

Note: Data are mean ± standard deviation. For diagnostic accuracy, Cohen's κ coefficients are given; reporting time is given in seconds [s]. Type-III Wald Chi-square tests and post-hoc tests were used for statistical modeling, and P-values highlighted in bold indicate significant differences against the threshold of α≤.01. The neural network's standalone diagnostic accuracy versus the majority vote of six expert radiologists was κ=0.81.

FT – free-text reporting; SR – structured reporting; AI-SR – AI-prefilled structured reporting.

**Table 4: Eye-Tracking Metrics per Radiography as a Function of Reader Experience, Area of Interest (AOI), and Reporting Mode.**

| Eye-Tracking-Metric | Reader Experience | Area of Interest | FT | SR | AI-SR | P-value |
|---|---|---|---|---|---|---|
| Total Fixation Duration [s] | All | Report | 11.4 ± 4.7 | 4.8 ± 2.6 | 3.6 ± 0.8 | **<.001** |
| | All | Radiograph | 18.2 ± 13.2 | 11.3 ± 8.9 | 7.6 ± 4.1 | .008 |
| | Low | Report | 13.6 ± 5.4 | 5.8 ± 3.5 | 3.1 ± 0.4 | **<.001** |
| | Low | Radiograph | 28.4 ± 10.4 | 16.5 ± 9.9 | 8.8 ± 5.4 | **<.001** |
| | High | Report | 9.2 ± 3.0 | 3.9 ± 0.8 | 4.2 ± 0.7 | **<.001** |
| | High | Radiograph | 8.1 ± 4.4 | 6.1 ± 3.6 | 6.5 ± 2.7 | .506 |
| Fixation Count [n] | All | Report | 37.5 ± 11.6 | 24.0 ± 13.5 | 19.0 ± 2.9 | **<.001** |
| | All | Radiograph | 72.4 ± 49.9 | 53.0 ± 43.7 | 34.0 ± 16.2 | .0098 |
| | Low | Report | 45.9 ± 10.2 | 31.0 ± 16.4 | 18.0 ± 2.3 | .0013 |
| | Low | Radiograph | 112.9 ± 35.8 | 80.0 ± 47.7 | 40.0 ± 20.4 | **<.001** |
| | High | Report | 29.1 ± 4.7 | 17.0 ± 3.5 | 21.0 ± 2.6 | **<.001** |
| | High | Radiograph | 32.0 ± 12.7 | 26.0 ± 16.6 | 28.0 ± 10.0 | .441 |
| Saccade Count [n] | All | Report | 99.9 ± 40.5 | 52.6 ± 18.7 | 54.1 ± 22.0 | **<.001** |
| | All | Radiograph | 205.1 ± 134.8 | 123.1 ± 88.3 | 96.9 ± 58.0 | **<.001** |
| | Low | Report | 114.1 ± 28.2 | 65.9 ± 16.6 | 53.1 ± 21.0 | **<.001** |
| | Low | Radiograph | 290.5 ± 125.5 | 182.5 ± 90.2 | 122.6 ± 57.8 | **<.001** |
| | High | Report | 85.7 ± 45.7 | 39.4 ± 8.2 | 55.0 ± 22.8 | **<.001** |
| | High | Radiograph | 119.7 ± 77.6 | 63.7 ± 20.2 | 71.2 ± 45.3 | .092 |
| Total Visit Duration [s] | All | Report | 16.9 ± 5.2 | 6.0 ± 3.2 | 4.6 ± 0.8 | **<.001** |
| | All | Radiograph | 25.1 ± 17.2 | 16.5 ± 13.9 | 10.6 ± 5.7 | **<.001** |
| | Low | Report | 19.1 ± 6.4 | 7.5 ± 4.1 | 4.0 ± 0.5 | **<.001** |
| | Low | Radiograph | 38.4 ± 14.4 | 25.5 ± 14.6 | 12.9 ± 7.1 | **<.001** |
| | High | Report | 14.8 ± 3.1 | 4.5 ± 0.9 | 5.2 ± 0.6 | **<.001** |
| | High | Radiograph | 11.9 ± 4.2 | 7.5 ± 4.8 | 8.3 ± 3.1 | .021 |
| Visit Count [n] | All | Report | 10.0 ± 2.3 | 8.1 ± 4.4 | 6.2 ± 1.6 | **.003** |
| | All | Radiograph | 9.1 ± 2.9 | 8.8 ± 5.8 | 5.9 ± 1.6 | .012 |

|  | Low | Report | 11.5 ± 1.95 | 10.8 ± 4.9 | 6.1 ± 1.8 | **.0014** |
|  | Low | Radiograph | 10.8 ± 2.94 | 12.5 ± 6.2 | 6.1 ± 1.9 | **<.001** |
|  | High | Report | 8.5 ± 1.74 | 5.4 ± 1.3 | 6.3 ± 1.5 | **.003** |
|  | High | Radiograph | 7.4 ± 1.75 | 5.0 ± 1.4 | 5.6 ± 1.5 | .018 |

Note: Data are mean ± standard deviation. Type-III Wald Chi-square tests were used for statistical modeling, and P-values highlighted in bold indicate significant differences against the significance threshold of α≤.01. Detailed post-hoc tests are provided in **Supplementary Table 4**. Abbreviations as in **Table 3**.

The distributions of visual attention across the radiograph varied depending on reader experience and reporting mode, as indicated by heatmap overlays of fixations on both display fields (**Figure 4**). Structured reporting and AI-prefilled structured reporting led to more clustered visual attention in the report display field, which was more diffuse for free-text reporting. Despite individual differences, all readers concentrated on the key anatomic regions required by the itemized findings in structured reporting and AI-prefilled structured reporting. Consequently, the lung apices and extrapulmonary and extracardiac structures tended to receive less attention. Structured reporting and AI-prefilled structured reporting resulted in clustered and intense visual attention on the central and basal lung fields.

The reporting mode was more relevant to diagnostic accuracy and efficiency than experience level (**Supplementary Table 6**). For eye-tracking metrics, experience level was a stronger predictor for fixation-based metrics and reporting mode for visit-based metrics. The AOI had a significant impact on most eye-tracking metrics.

**User experience** surveys indicated that structured reporting and AI-prefilled structured reporting modes were more favored regarding user satisfaction, user interface convenience, perceived efficiency, and willingness for clinical adoption (**Supplementary Table 7)**. While the usefulness of AI was generally rated as 'high,' trust in the AI suggestions remained limited. The usage of structured reporting and AI-prefilled structured reporting reduced the (subjective) cognitive burden relative to free-text reporting.

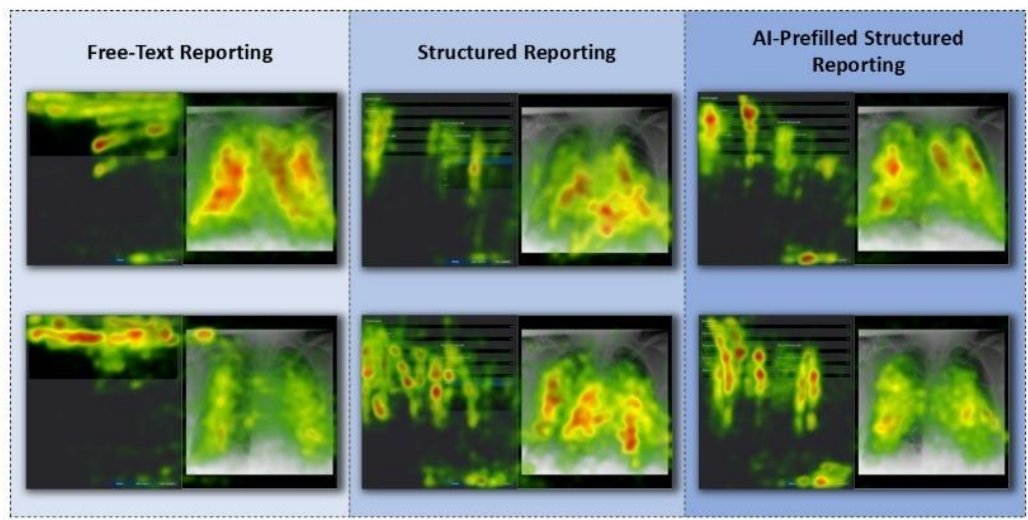

**Figure 4**: **Heatmap Overlays of Fixations as a Function of Reader Experience and Reporting Mode.** Each panel shows fixation heatmaps from all readers of the corresponding experience group combined (rows) across the three reporting modes (columns). Within each panel, the radiograph display field is located on the right, and the corresponding report display field on the left. The heatmaps illustrate the spatial distribution of cumulative fixation durations, with green indicating lower and red higher fixation intensity. Color scaling was applied individually to each heatmap, meaning fixation intensities are not directly comparable across panels. Heatmaps were generated in the post-processing stage using Tobii Studio.

## 4. Discussion

This study evaluated how different reporting modes —free-text, structured, and AI-prefilled structured reporting— affect radiologists' diagnostic accuracy, efficiency, image analysis behavior, and user experience when reporting on bedside chest radiographs. We found that diagnostic accuracy did not differ between free-text and structured reporting but increased with AI assistance, independent of experience level. With and without AI assistance, structured reporting focused visual attention and streamlined visual processing, especially among novice readers. For these readers, structured reporting also shifted visual attention from the report to the radiograph. Conversely, non-novice readers maintained a steady focus on the radiograph, irrespective of the reporting mode.

In radiology, structured reporting is a promising yet controversial innovation aimed at enhancing the clarity, consistency, and utility of radiologic reports (19, 20). However, its impact on visual search patterns is unclear. Comparing structured reporting and free-text reporting based on five brain MRI studies, Harris et al. provided anecdotal evidence that structured reporting decreases diagnostic efficiency and diverts the reader's focus from the image to the report (21). This observation contrasts with our results, which indicate increased diagnostic efficiency and consistent prioritization of visual attention toward the radiograph over the report, irrespective of reader experience. These contradictory findings may be due to modality-specific and methodologic differences, as Harris et al. had free-text reports created by dictation, while we only allowed typed reports.

In our study, improved diagnostic efficiency was associated with readers spending less time processing the radiograph and report, suggesting more efficient information extraction and reduced cognitive strain, as indicated by self-reported user experience surveys. Similarly, with structured reporting, eye-tracking metrics were quantitatively lower, which –together with the cumulative heatmaps– reflects more efficient search strategies focusing on key anatomic regions. Moreover, the maintenance of total fixation and visit duration on the radiograph suggests that readers continued to devote sufficient time to analyzing the image. Structured reporting appears to offer operational benefits by improving diagnostic efficiency without compromising accuracy or thorough visual analyses. However, it is also evident that structured reporting –as administered in our study– reduces the need to explore the image extensively, as non-reported structures like the lung apices or extrapulmonary structures received less visual attention.

Integrating AI assistance into the diagnostic process improved diagnostic accuracy and efficiency, reinforcing its role as a promising decision-support tool (22). Consistent with prior research (23-25), we observed significant improvements in diagnostic performance across all readers who used AI assistance. By providing preliminary readings, our previously validated AI model helped non-radiologist physicians reach near-expert radiologist performance (12) and, in the present study, supported readers in identifying critical findings. Novice readers benefited most from AI assistance, elevating their diagnostic accuracy to that of non-novice readers. Interestingly, non-novice readers improved efficiency while consistently engaging with the radiograph across all reporting modes, indicating their ability to utilize AI assistance without compromising their established visual search strategies. In contrast, novice readers relied more on the visual guidance provided by the structured template, which may help minimize distractions associated with report generation but may predispose to automation bias and overreliance (9). Although AI assistance improved accuracy compared with free-text and structured reporting, it still fell short of the model's standalone performance. This pattern resembles the well-described phenomenon of algorithmic aversion, in which radiologists sometimes down-weight or override even correct AI suggestions when the system is framed as experimental (26). Consistent with this notion, six of eight readers reported low trust in the AI in our user experience survey.

This study has limitations. First, the small sample size of 35 radiographs and the focus on chest radiography limit generalizability. Second, the experimental setting favored scientific standardization but did not fully emulate clinical practice. Therefore, future studies should use eye trackers that integrate seamlessly into the workflow, and actual radiology workstations with multiple screens, provide access to prior imaging studies and clinical information, and allow for image manipulation functions such as zooming or panning. Third, we required study participants to report using standardized text snippets while voice dictation was disabled. Although this differs from dictation-centered workflows elsewhere, it mirrors our local routine in which chest-radiograph reports are typed using standardized snippets. It may also influence visual attention patterns. Presumably, a radiologist's gaze would be more directed toward the report display field when modifying a text snippet, as would occur during voice dictation. Nonetheless, manual text editing (secondary to structured reporting or dictation) is required, regardless of the input format, and remains unaccounted for in this study. Fourth, because all participants completed the three reading sessions in a fixed order (free-text reporting → structured reporting → AI-prefilled structured reporting) using the same radiographs each session, residual learning or order effects cannot be fully excluded. However, the two-week wash-out period, the absence of feedback, the abrupt (rather than gradual) shift in efficiency and gaze metrics between free-text and structured reporting, and the unchanged diagnostic accuracy between free-text and structured reporting argue against a relevant recall bias. Performance improved only when the AI model predictions were added in the final session, further supporting the notion of true reporting mode-associated effects. Fifth, the template used for structured reporting did not include free-text sections for reporting findings like pneumothorax, soft tissue, bone, or other abnormalities, unlike the template used clinically, which is administered sequentially in multiple frames; thus, efficiency gains may be due to a stronger focus on template-queried items. Sixth, we used a well-validated yet academic AI model without FDA or CE certification. The model was trained and validated on a large, heterogeneous dataset with strict separation of training, validation, and temporally independent test sets (see **Supplementary Text 1**). While such measures reduce the risk of dataset-specific bias, they cannot replace independent multi-center validation across hospitals, scanners, and patient populations. Future work should incorporate explicit bias-control strategies, emphasize transparency and explainability, and ultimately confirm robustness through multi-center validation and prospective trials before clinical adoption. Finally, our readers were residents or pre-graduate medical students, limiting the applicability of our conclusions to more experienced radiologists. In addition to addressing the mentioned limitations, future studies should include the analysis of granular gaze metrics, such as time-to-first-fixation (on a finding) or the analysis of short, uncommitted glances, which could further elucidate search and recognition errors. However, these studies require single-finding radiographs with pixel-level annotations; the multi-finding radiographs used here do not permit such analyses. This analysis may shed light on non-conscious processes during the visual assessment of imaging studies (27), helping to reduce failures of conscious recognition and, potentially, diagnostic errors. Interestingly, the standalone AI model outperformed AI-assisted readers. Beyond algorithmic accuracy, the AI output timing (concurrent vs second-reader workflows), its framing (e.g., experimental vs validated), and the design of user interfaces may critically influence how radiologists adopt AI suggestions. While our study did not systematically assess these factors, they represent important directions for future research to optimize human-AI collaboration.

In conclusion, different reporting modes significantly impact diagnostic accuracy and efficiency. Structured reporting enhances efficiency by directing visual attention toward the image, particularly benefiting inexperienced readers. AI-prefilled structured reporting improves diagnostic accuracy.

## Funding information

This research is supported by the Deutsche Forschungsgemeinschaft - DFG (701010997, 517243167, 515639690) , the German Federal Ministry of Research, Technology and Space (Transform Liver - 031L0312C, DECIPHER-M, 01KD2420B) and the European Union Research and Innovation Programme (ODELIA - GA 101057091, SAGMA – GA 101222556).

## Author Contributions

R.S. and S.N. conceived and designed the study. The manuscript was written by M.K. and reviewed, critically edited, and refined by R.S. and S.N. Methodology was developed by R.S. and M.K. Software implementation was carried out by M.K. The experimental investigation was performed by M.K. and R.S. Data curation and preprocessing were conducted by M.K., while data interpretation and analysis were jointly performed by M.K. and R.S. Formal statistical analysis was undertaken by M.K. and R.S., and visualization was prepared by M.K. and R.S. R.S. supervised the study and managed the project. Clinical expertise and interpretation were provided by M.vdS., F.B.O., A.I., E.N.C., J.N., A.K., V.P., A.S., D.PdS., C.K., D.T., S.N., and R.S. All authors approved the final version of the manuscript and agree to be accountable for all aspects of the work.

# Supplementary Material

**Supplementary Table 1: Frequencies of Itemized Imaging Findings as Referenced by the Majority Vote of Six Expert Radiologists.**

| Item | Quantity n (%) | |
|---|---|---|
| Cardiomegaly | | |
|   Normal | 14 (40) | |
|   Borderline | 3 (8.6) | |
|   Enlarged | 14 (40) | |
|   Massively Enlarged | 2 (5.7) | |
|   Not assessable | 1 (2.9) | |
| Pleural effusion | Left | Right |
|   None | 20 (57.1) | 24 (68.6) |
|   (+) | 0 (0) | 0 (0) |
|   + | 12 (34.3) | 8 (22.9) |
|   ++ | 3 (8.6) | 3 (8.6) |
|   +++ | 0 (0) | 0 (0) |
| Pulmonary Opacities (left or right) | Left | Right |
|   None | 30 (85.7) | 27 (77.1) |
|   (+) | 0 (0) | 0 (0) |
|   + | 1 (2.9) | 4 (11.4) |
|   ++ | 4 (11.4) | 4 (11.4) |
|   +++ | 0 (0) | 0 (0) |
| Pulmonary Congestion | | |
|   None | 18 (51.4) | |
|   (+) | 2 (5.7) | |
|   + | 13 (37.1) | |
|   ++ | 2 (5.7) | |
|   +++ | 0 (0) | |
| Atelectasis (left or right) | Left | Right |
|   None | 18 (51.4) | 15 (42.9) |
|   (+) | 0 (0) | 0 (0) |
|   + | 14 (0.4) | 19 (54.3) |
|   ++ | 3 (8.6) | 1 (2.9) |
|   +++ | 0 (0) | 0 (0) |

**Supplementary Table 2: Post-Session Questionnaire.**

Each reader was asked to provide structured feedback at the end of each session.

| Dimension of User Feedback | Question | Response Scale |
|---|---|---|
| **Diagnostic Confidence** | How confident are you in your diagnoses using the reporting mode? | low = 'very insecure' <br> neutral = 'moderately confident' <br> high = 'very confident' |
| **User Satisfaction Level** | How satisfied are you with the reporting mode? | low = 'very dissatisfied' <br> neutral = 'indifferent' <br> high = 'very satisfied' |
| **Cognitive Ease** | How much cognitive ease does the reporting mode provide? | low = 'low cognitive ease' (= high cognitive effort) <br> neutral = 'moderate cognitive ease' (= moderate cognitive effort) <br> high = 'high cognitive ease' (= low cognitive effort) |
| **User Interface – Convenience** | How convenient is it to use the reporting mode? | low = 'very inconvenient' <br> neutral = 'moderately convenient' <br> high = 'very convenient' |
| **User Interface – Efficiency** | How efficient is the reporting mode? | low = 'very inefficient' <br> neutral = 'moderately efficient' <br> high = 'very efficient' |
| **Clinical Adoption Willingness** | How willing are you to use this reporting mode in your clinical routine? | low = 'by no means' <br> neutral = 'indifferent' <br> high = 'by all means' |
| **Usefulness of AI Assistance (*)** | How useful do you find the AI assistance? | low = 'not useful' <br> neutral = 'moderately useful' <br> high = 'very useful' |
| **Trust in AI Assistance (*)** | To what extent do you trust the AI predictions? | low = 'not at all' <br> neutral = 'partially' <br> high = 'very much' |

Note: (*) 'Usefulness of AI Assistance' and 'Trust in AI Assistance' were only surveyed after the AI-assisted reading session (AI-SR).

**Supplementary Table 3: Additional Diagnostic Performance Measures as a Function of Reporting Mode and Reader Experience.**

| Reader Experience | Metric | FT | SR | AI-SR |
|---|---|---|---|---|
| All | TP [n] | 698 | 733 | 775 |
| All | FP [n] | 371 | 451 | 311 |
| All | TN [n] | 983 | 899 | 1039 |
| All | FN [n] | 137 | 107 | 89 |
| All | Sensitivity [%] | 84 | 87 | 90 |
| All | Specificity [%] | 73 | 67 | 77 |
| All | Accuracy [%] | 77 | 74 | 82 |
| All | Cohen's κ | 0.58 | 0.60 | 0.71 |
| Low | TP [n] | 336 | 380 | 392 |
| Low | FP [n] | 196 | 291 | 161 |
| Low | TN [n] | 481 | 383 | 516 |
| Low | FN [n] | 78 | 37 | 36 |
| Low | Sensitivity [%] | 81 | 91 | 92 |
| Low | Specificity [%] | 71 | 57 | 76 |
| Low | Accuracy [%] | 75 | 70 | 82 |
| Low | Cohen's κ | 0.54 | 0.54 | 0.71 |
| High | TP [n] | 362 | 353 | 383 |
| High | FP [n] | 175 | 160 | 150 |
| High | TN [n] | 502 | 516 | 523 |
| High | FN [n] | 59 | 70 | 53 |
| High | Sensitivity [%] | 86 | 84 | 88 |
| High | Specificity [%] | 74 | 76 | 78 |
| High | Accuracy [%] | 79 | 79 | 82 |
| High | Cohen's κ | 0.62 | 0.66 | 0.71 |

Note: TP (true positive), FP (false positive), TN (true negative), and FN (false negative) are provided as cumulative counts for all readers (within each experience level or combined) and all radiographs read per session. Cohen's κ coefficients are presented as mean values. Abbreviations: FT – free-text reporting; SR – structured reporting; AI-SR – AI-prefilled structured reporting.

**Supplementary Table 4: Post-Hoc Details of Eye-Tracking Metrics as a Function of Reporting Mode, Area of Interest (AOI), and Reader Experience.**

| Eye-Tracking Metric | Reader Experience | Area of Interest | FT vs. SR | FT vs. AI-SR | SR vs. AI-SR | FT Radiograph vs. Report | SR Radiograph vs. Report | AI-SR Radiograph vs. Report |
|---|---|---|---|---|---|---|---|---|
| Total Fixation Duration | All | Both | * | ** | ns | ns | * | * |
| | Low | Both | ns | * | ns | ns | * | * |
| | High | Both | ns | ns | ns | ns | ns | ns |
| | All | Report | ** | ** | ns | | | |
| | All | Radiograph | ns | ns | ns | | | |
| | Low | Report | ns | * | ns | | | |
| | Low | Radiograph | ns | * | ns | | | |
| | High | Report | ** | ** | ns | | | |
| | High | Radiograph | ns | ns | ns | | | |
| Fixation Count | All | Both | ns | * | ns | * | * | ns |
| | Low | Both | ns | * | ns | * | * | * |
| | High | Both | ns | ns | ns | ns | ns | ns |
| | All | Report | ns | * | ns | | | |
| | All | Radiograph | ns | ns | ns | | | |
| | Low | Report | ns | ns | ns | | | |
| | Low | Radiograph | ns | * | ns | | | |
| | High | Report | ns | * | ns | | | |
| | High | Radiograph | ns | ns | ns | | | |
| Saccade Count | All | Both | * | ** | ns | * | ** | * |
| | Low | Both | ns | ** | ns | ** | ** | ** |
| | High | Both | ns | ns | ns | ns | ns | ns |
| | All | Report | * | * | ns | | | |
| | All | Radiograph | ns | ** | ns | | | |
| | Low | Report | ns | ns | ns | | | |
| | Low | Radiograph | * | ** | ns | | | |
| | High | Report | * | * | ns | | | |
| | High | Radiograph | ns | ns | ns | | | |
| Total Visit Duration | All | Both | | | | ns | ** | * |
| | Low | Both | | | | ns | ** | ** |
| | High | Both | | | | ns | ns | ns |
| | All | Report | ** | ** | ns | | | |

|  | All | Radiograph | ns | * | ns |  |  |  |
|---|---|---|---|---|---|---|---|---|
|  | Low | Report | ns | * | ns |  |  |  |
|  | Low | Radiograph | ns | * | ns |  |  |  |
|  | High | Report | ** | ** | ns |  |  |  |
|  | High | Radiograph | ns | ns | ns |  |  |  |
| **Visit Count** | All | Both |  |  |  | ns | ns | ns |
|  | Low | Both |  |  |  | ns | ns | ns |
|  | High | Both |  |  |  | ns | ns | ns |
|  | All | Report | ns | * | ns |  |  |  |
|  | All | Radiograph | ns | ns | ns |  |  |  |
|  | Low | Report | ns | ns | ns |  |  |  |
|  | Low | Radiograph | ns | ns | ns |  |  |  |
|  | High | Report | ns | * | ns |  |  |  |
|  | High | Radiograph | ns | ns | ns |  |  |  |

Note: ns – not significant; * – significant (.001<P≤.01); ** – very significant (P≤.001). For each eye-tracking metric, post-hoc details are organized as follows: Reporting modes are compared for both display fields combined as a function of reader experience (highlighted in grey). Next, for a specific reporting mode, the two display fields, radiograph and report, are compared as a function of reader experience (highlighted in light grey). Last, for a specific display field, the reporting modes are compared pairwisely and as a function of reader experience (highlighted in white).

**Supplementary Table 5: Relative Changes in Eye-Tracking Metrics as a Function of Reader Experience and Area of Interest (AOI).**

| Eye-Tracking Metric | Reader Experience | Area of Interest | % Change SR vs. FT | % Change AI-SR vs FT | % Change AI-SR vs SR |
|---|---|---|---|---|---|
| Total Fixation Duration | All | Report | -58 | -68 | -25 |
| | All | Radiograph | -38 | -58 | -33 |
| | Low | Report | -57 | -77 | -47 |
| | Low | Radiograph | -42 | -69 | -47 |
| | High | Report | -58 | -54 | +8 |
| | High | Radiograph | -25 | -20 | +7 |
| Fixation Count | All | Report | -36 | -49 | -21 |
| | All | Radiograph | -27 | -53 | -36 |
| | Low | Report | -32 | -61 | -42 |
| | Low | Radiograph | -29 | -65 | -50 |
| | High | Report | -42 | -28 | +24 |
| | High | Radiograph | -19 | -13 | +8 |
| Saccade Count | All | Report | -47 | -46 | +3 |
| | All | Radiograph | -40 | -53 | -21 |
| | Low | Report | -42 | -53 | -19 |
| | Low | Radiograph | -37 | -58 | -33 |
| | High | Report | -54 | -36 | +40 |
| | High | Radiograph | -47 | -41 | +12 |
| Total Visit Duration | All | Report | -64 | -73 | -23 |
| | All | Radiograph | -34 | -58 | -36 |
| | Low | Report | -61 | -79 | -47 |
| | Low | Radiograph | -34 | -66 | -49 |
| | High | Report | -70 | -65 | +16 |
| | High | Radiograph | -37 | -30 | +11 |
| Visit Count | All | Report | -19 | -38 | -23 |
| | All | Radiograph | -3 | -35 | -33 |
| | Low | Report | -6 | -47 | -44 |
| | Low | Radiograph | +16 | -44 | -51 |
| | High | Report | -36 | -26 | +17 |
| | High | Radiograph | -32 | -24 | +12 |

Note: Abbreviations as in Table 2.

**Supplementary Table 6: Quantification of Predictor Variables of the Generalized Linear Mixed Effects Model for Diagnostic Accuracy, Diagnostic Efficiency, and Eye-Tracking Metrics.**

| Outcome Parameter | Predictor | Coefficient [95% CI] | P-value |
|---|---|---|---|
| **Diagnostic Accuracy** | Experience_Level[Low] | -0.29 [-0.49; -0.10] | **.004** |
|  | Reporting_mode[SR] | 0.06 [-0.09; 0.22] | .433 |
|  | Reporting_mode[AI-SR] | 0.56 [0.39; 0.72] | **<.001** |
| **Diagnostic Efficiency** | Experience_Level[Low] | 0.58 [0.27; 0.88] | **<.001** |
|  | Reporting_mode[SR] | -0.93 [-1.06; -0.80] | **<.001** |
|  | Reporting_mode[AI-SR] | -1.29 [-1.42; -1.16] | **<.001** |
| **Total Fixation Duration** | Experience_Level[Low] | 0.83 [0.35; 1.31] | **<.001** |
|  | Reporting_mode[SR] | -0.59 [-1.03; -0.14] | **.010** |
|  | Reporting_mode[AI-SR] | -0.47 [-0.92; -0.03] | .038 |
|  | Area_of_Interest[Report] | -0.54 [-0.79; -0.28] | **<.001** |
| **Fixation Count** | Experience_Level[Low] | 0.86 [0.42; 1.30] | **<.001** |
|  | Reporting_mode[SR] | -0.42 [-0.78; -0.07] | .020 |
|  | Reporting_mode[AI-SR] | -0.23 [-0.58; 0.13] | .216 |
|  | Area_of_Interest[Report] | -0.52 [-0.72; -0.31] | **<.001** |
| **Saccade Count** | Experience_Level[Low] | 0.70 [0.13; 1.27] | .015 |
|  | Reporting_mode[SR] | -0.56 [-0.89; -0.23] | **<.001** |
|  | Reporting_mode[AI-SR] | -0.42 [-0.75; -0.09] | .012 |
|  | Area_of_Interest[Report] | -0.57 [-0.76; -0.38] | **<.001** |
| **Total Visit Duration** | Experience_Level[Low] | 0.69 [0.24; 1.14] | **.003** |
|  | Reporting_mode[SR] | -0.88 [-1.30; -0.46] | **<.001** |
|  | Reporting_mode[AI-SR] | -0.70 [-1.12; -0.28] | **.001** |
|  | Area_of_Interest[Report] | -0.58 [-0.83; -0.34] | **<.001** |
| **Visit Count** | Experience_Level[Low] | 0.32 [-0.02; 0.67] | .067 |
|  | Reporting_mode[SR] | -0.45 [-0.66; -0.23] | **<.001** |
|  | Reporting_mode[AI-SR] | -0.31 [-0.53; -0.09] | **.005** |
|  | Area_of_Interest[Report] | 0.05 [-0.07; 0.18] | .425 |

Note: The coefficients and p-values for each predictor are presented, quantifying their effects on diagnostic accuracy, diagnostic efficiency, and eye-tracking metrics. The coefficients represent the change in the log values of the corresponding outcome metric compared to the reference, i.e., high (experience level), free-text (reporting mode), and report display field (Area of Interest). Coefficients are given as Mean [95 % Confidence Interval]. Statistically significant associations are indicated in bold type. Read the table as follows: For each outcome measure listed in the first column, coefficients indicate how that measure changes on the log scale when the respective predictor (e.g., low reader experience) is compared to its reference category (e.g., high reader experience), with positive values indicating an increase and negative values a decrease. **Bold** font indicates statistically significant p-values (p < 0.05).

**Supplementary Table 7: Results of User Experience Surveys as a Function of Reporting Mode and Reader Experience.**

| Survey Dimension | FT | SR | AI-SR |
|---|---|---|---|
| **Diagnostic confidence** | | | |
|     All Readers | 0/2/6 | 0/4/4 | 1/2/5 |
|     Novice Readers | 0/2/2 | 0/2/2 | 1/1/2 |
|     Non-novice Readers | 0/0/4 | 0/2/2 | 0/1/3 |
| **User satisfaction level** | | | |
|     All Readers | 2/3/3 | 1/0/7 | 0/0/8 |
|     Novice Readers | 0/3/1 | 1/0/3 | 0/0/4 |
|     Non-novice Readers | 2/0/2 | 0/0/4 | 0/0/4 |
| **Cognitive ease** | | | |
|     All Readers | 5/2/1 | 2/4/2 | 0/6/2 |
|     Novice Readers | 2/1/1 | 1/3/0 | 0/4/0 |
|     Non-novice Readers | 3/1/0 | 1/1/2 | 0/2/2 |
| **User interface – convenience** | | | |
|     All Readers | 1/3/4 | 0/1/7 | 0/1/7 |
|     Novice Readers | 1/1/2 | 0/0/4 | 0/1/3 |
|     Non-novice Readers | 0/2/2 | 0/1/3 | 0/0/4 |
| **User interface – perceived efficiency** | | | |
|     All Readers | 3/4/1 | 0/0/8 | 0/1/7 |
|     Novice Readers | 0/3/1 | 0/0/4 | 0/1/3 |
|     Non-novice Readers | 3/1/0 | 0/0/4 | 0/0/4 |
| **Clinical adoption willingness** | | | |
|     All Readers | 4/3/1 | 0/0/8 | 0/1/7 |
|     Novice Readers | 1/2/1 | 0/0/4 | 0/1/3 |
|     Non-novice Readers | 3/1/0 | 0/0/4 | 0/0/4 |
| **Usefulness of AI assistance (*)** | | | |
|     All Readers | - | - | 0/2/6 |
|     Novice Readers | - | - | 0/2/2 |
|     Non-novice Readers | - | - | 0/0/4 |
| **Trust in AI assistance (*)** | | | |
|     All Readers | - | - | 6/1/1 |
|     Novice Readers | - | - | 3/1/0 |
|     Non-novice Readers | - | - | 3/0/1 |

Note: Users indicated their experience by selecting low, neutral, or high for the respective survey dimension. Data are presented as counts of low/neutral/high. Abbreviations as in Table 2. (*) 'Usefulness of AI Assistance' and 'Trust in AI Assistance' were only surveyed after the AI-assisted reading session (AI-SR).

**Supplementary Text 1: Details of the neural network.**

We utilized the AI model created by Khader et al. (1) to aid in the interpretation of bedside chest radiographs. Tailored specifically for intensive care unit (ICU) environments, this model was trained on a dataset of 193,566 radiographs collected over a 12-year time period (01/2009 – 12/2020). The dataset also incorporated structured, semiquantitative radiologic reports written during routine reporting and used for clinical communication. Unlike other models that rely on automatic label extraction procedures for post-hoc dataset annotation, our dataset was collected prospectively. Trained radiologists provided machine-readable labels at the time of reporting, while they were aware of a patient's clinical situation, history, and other imaging and non-imaging findings. This dataset was used to train an AI model that provided predictions on the presence or absence, as well as the severity (or distribution) of five critical findings: cardiomegaly, pulmonary congestion, pleural effusion, pulmonary opacities, and atelectasis.

More specifically, a ResNet-34 convolutional neural network, specifically designed for multi-label classification, was used. Each finding reflects severity gradings that range from none to severe, ensuring that the classifications for each imaging finding are mutually exclusive. The radiographs underwent processing through a Sigmoid activation layer to generate probability scores. Final predictions were determined by selecting the classification with the highest probability. For training purposes, 122,294 radiographs were used, with 31,243 reserved for validation. The training utilized the Adam optimizer in PyTorch and lasted for 100 epochs, approximately five days on an Nvidia Quadro RTX 6000. The full code is also accessible on GitHub (https://github.com/FirasGit/chest_radiography_ai_vs_hi). An additional internal test set of 40,029 radiographs from 9,004 patients was held out and used only for evaluation, ensuring a clear separation of training, validation, and testing cohorts.

The whole dataset was highly heterogeneous, encompassing more than 45,000 patients, 18 different radiography systems, and images acquired over a decade of routine clinical practice. It included examinations from 10 different intensive care units covering a great number of clinical disciplines and was acquired by more than 70 trained radiologic technologists. Inclusion of a large range of acquisition devices, patient populations, clinical settings, and operators aimed for generalizability beyond a narrowly defined dataset.

In addition to the internal test set, a temporally separated hold-out set of 100 radiographs from 100 patients acquired in 2021 was employed. The AI model achieved expert-level agreement with the majority vote of six expert radiologists (Cohen's κ = 0.86). Given this validated performance, we incorporated the model as a clinical decision-support tool for interpreting bedside chest radiographs in this study.

**Supplementary Text 2: Details of Power Analysis**

Guided by a meta-analysis on eye-tracking measures in medicine and beyond ((27), r=0.29, moderate effect size), we used G*Power (v3.1.9.7; Heinrich-Heine-University, Duesseldorf, Germany) (28) to determine the minimum sample size, i.e., the minimum number of radiographs per session. We assumed an α error of 0.05, a power of 0.80, a moderate correlation coefficient (among repeated measures) ρ of 0.5, a nonsphericity correction of 0.75, a moderate effect size f of 0.25 (29), and a repeated-measures design with three sessions ("F tests," "ANOVA: Repeated measures, within factors," "number of groups": 1 [within-subject], "number of measurements": 3 [number of reporting modes]). The minimum sample size thus determined was 34. For improved power, we included 35 radiographs in the analysis.

**Supplementary Text 3: Calculation of the saccade count using Tobii raw data.**

Tobii Studio uses the I-VT fixation classifier as a default setting for detecting fixations and saccades. The I-VT fixation classifier, based on the Velocity-Threshold Identification (I-VT) filter described by Salvucci and Goldberg (1) and Komogortsev et al. (2), categorizes eye-tracking data into fixations and saccades by applying an angular velocity threshold. Each data point is assigned an angular velocity, which is calculated using Tobii Studio's Velocity Calculator. Data points with angular velocities below the threshold (default: 30°/s) are classified as fixations, while those above are classified as saccades. This method clearly distinguishes between fixation and saccade events based on velocity.

The raw data includes the variables 'StrictAverageGazePointX' and 'StrictAverageGazePointY', representing the horizontal and vertical coordinates (in millimeters) of the averaged gaze points for both eyes on the screen during each saccade. The number of saccades was calculated by filtering the raw data (as exported from the Tobii platform) based on two key parameters: 'GazeEventType' set to 'Saccade' and 'SaccadeIndex', allowing the determination of the average number of saccades per image (study). AOIs were defined by further filtering the 'StrictAverageGazePointX' and 'StrictAverageGazePointY' columns, restricting gaze points to a specific region within the AOIs, measured in millimeters.